\documentclass[journal]{IEEEtran}

\ifCLASSINFOpdf
\else
   \usepackage[dvips]{graphicx}
\fi
\usepackage{url}

\usepackage{graphicx}
\usepackage{cite}
\usepackage{bm}
\usepackage{enumitem}
\usepackage{algorithm}
\usepackage{algorithmic}
\usepackage{amssymb}
\usepackage{amsmath}
\usepackage{multirow}
\usepackage{color}

\begin{document}

\title{DropDim: A Regularization Method for Transformer Networks}

\author{Hao Zhang, Dan Qu, Keji Shao, and Xukui Yang
\thanks{Manuscript received September 21, 2021; revised December 20, 2021; accepted December 25, 2021. Date of publication January 5, 2022; date of current version February 1, 2022. This work was supported by the National Natural Science Foundation of China under Grants 61673395 and 62171470. The associate editor coordinating the review of this manuscript and approving it for publication was Dr. Odette Scharenborg. \textit{(Corresponding authors: Dan Qu; Xukui Yang.)}}
\thanks{Hao Zhang, Dan Qu, Xukui Yang are with the School of the Information Engineering University, Zhengzhou 450000, China.(e-mail:haozhang0126@163.com;qudanqudan@sina.com;gzyangxk@163.com).}
\thanks{Keji Shao is with Jiangnan Institute of Computing Technology. (e-mail:shaokjppl@163.com).

Digital Object Identifier 10.1109/LSP.2022.3140693}}

\markboth{IEEE SIGNAL PROCESSING LETTERS, Vol. 29, 2022}
{Shell \MakeLowercase{\textit{et al.}}: Bare Demo of IEEEtran.cls for IEEE Journals}
\maketitle

\begin{abstract}
We introduce DropDim, a structured dropout method designed for regularizing the self-attention mechanism, which is a key component of the transformer. In contrast to the general dropout method, which randomly drops neurons, DropDim drops part of the embedding dimensions. In this way, the semantic information can be completely discarded. Thus, the excessive co-adapting between different embedding dimensions can be broken, and the self-attention is forced to encode meaningful features with a certain number of embedding dimensions erased. Experiments on a wide range of tasks executed on the MUST-C English-Germany dataset show that DropDim can effectively improve model performance, reduce over-fitting, and show complementary effects with other regularization methods. When combined with label smoothing, the WER can be reduced from 19.1\% to 15.1\% on the ASR task, and the BLEU value can be increased from 26.90 to 28.38 on the MT task. On the ST task, the model can reach a BLEU score of 22.99, an increase by 1.86 BLEU points compared to the strong baseline.
\end{abstract}


\IEEEpeerreviewmaketitle

\section{Introduction}

\IEEEPARstart{T}{he} end-to-end (E2E) neural network has achieved great success in sequence-to-sequence problems such as Automatic Speech Recognition (ASR) \cite{chan2016listen}, Machine Translation (MT) \cite{sutskever2014sequence}, and Speech Translation (ST) \cite{weiss2017sequence}. In this paradigm, a single neural network is used to directly map the input to the target, removing the independent part of the cascade method. Transformer \cite{vaswani2017attention} is currently the most popular E2E structure and has achieved state of the art performance in many sequences modeling tasks \cite{dong2018speech,gangi2019adapting}. This ability is partly because it makes few assumptions about structural information of data, which makes the transformer a universal and flexible architecture \cite{lin2021a}. As a side effect, the lack of structural bias makes the transformer prone to overfitting for small-scale data.

Dropout \cite{srivastava2014dropout} is a widely used regularization strategy, which multiplies each neuron with a sample of a Bernoulli random variable during training, with probability $p$ of dropping/zeroing out the neuron. The final model can be understood as the average of multiple models. Several variations of dropout have also been proposed, such as dropblock \cite{ghiasi2018dropblock}, spatial dropout \cite{tompson2015efficient}, and zoneout \cite{krueger2016zoneout}. While most transformer-based models still use the regular dropout \cite{baevski2020wav2vec,zhang2020transformer}. 

DropAttention is recommended in \cite{zehui2019dropattention}. By randomly setting attention weights to zero, the final model was encouraged to use the full context of the input sequences for prediction, rather than relying solely on a small piece of features. And considering that the multi-head attention mechanism is dominated by a small number of attention heads \cite{michel2019are}, Zhou et al. \cite{zhou2020scheduled} recommends DropHead, which randomly discards part of the attention heads during the training.
 
The above methods directly process the attention head. In this paper, we recommend another simple structured dropout method from the perspective of self-attention layer input, called DropDim. The key motivation behind DropDim is that each embedding dimension contains certain semantic information in transformer. In other words, the basic unit is a vector instead of a single neuron \cite{zehui2019dropattention}. Thus, the independence assumption in dropout needs to be relaxed. We achieve this by dropping the embedding dimension. In this way, semantic information can be completely removed. The excessive co-adapting between different embedding dimensions can be broken, and the self-attention is forced to encode meaningful features with a certain number of embedding dimensions erased.

There exists some structured dropout like the one suggested by us \cite{wang2021specaugment1, xie2017dnsnnlm, gal2016theoretically, dalmia2021transformer, dalmia2021searchable}.The most similar to our method is cutoff \cite{shen2020a}, which is a set of simple yet efficient data augmentation strategies, including token cutoff, feature cutoff, and span cutoff. However, the motivation and application location are very different. Cutoff processes the input sentence to generate a restricted perspective, which helps enrich empirical observations and better cover the data space. While we process the input of the attention mechanism to ensure that self-attention layer can encode more meaningful features without utilizing the information from the removed embedding dimensions at all.

Experiments on a wide range of tasks executed on the MUST-C English-Germany dataset \cite{gangi2019must} show that DropDim can effectively improve model performance, reduce over-fitting, and show complementary effects with other regularization methods. When combined with label smoothing \cite{szegedy2016rethinking}, the WER can be reduced from 19.1\% to 15.1\% on the ASR task, and the BLEU value can be increased from 26.90 to 28.38 on the MT task. On the ST task, the model can reach a BLEU score of 22.99, an increase by 1.86 BLEU points compared to the strong baseline.

\section{Method}

DropDim is a simple regularization method similar to dropout. Its main difference from dropout is that it drops embedding dimensions with a certain probability instead of randomly dropping independent units.

\begin{equation}
\label{equa}
	\bm{h} = DropDim(\bm{x} + sub\_block(\bm{x}))
\end{equation}
Where $\bm{x}$ denotes the input of transformer encoder or decoder sub-block. Unless otherwise specified, DropDim is applied to all sub-blocks.

\begin{figure}
\centerline{\includegraphics[width=\columnwidth]{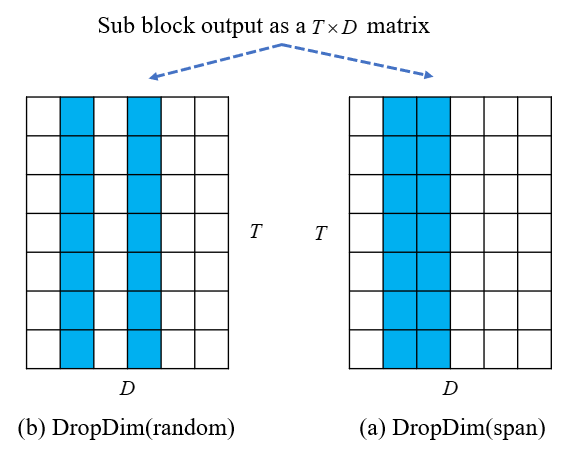}}
\caption{Schematic illustration of the proposed DropDim, including DropDim(random) and DropDim(span). The blue area indicates that the corresponding element is dropped. Similar to dropout we do not apply DropDim during inference.}
\label{fig:DropDim}
\end{figure}

In detail, we propose two structured dropout methods: DropDim(random) and DropDim(span), which are shown in Fig.~\ref{fig:DropDim}.
\begin{enumerate}
	\item DropDim(random) means dropping independent embedding dimension. Algorithm~\ref{algorithm1} describes its pseudo code.
	\item DropDim(span) means dropping span of the embedding dimension. Let $\alpha$ be the pre-defined value that determined the max length of the consecutive drop. Then, the span length $l$ is chosen from the uniform distribution on the interval [0, $\alpha$] and the starting index $s$ for the span is randomly sampled as: $s \in \{ 0,1,...,D - l\} $  with $D$ is the size of embedding dimension. Afterward, the embeddings between the  $s$-th and  $(s + l - 1)$-th positions are dropped.
\end{enumerate}

\begin{algorithm}[htb]
\caption{DropDim}
\begin{enumerate}
\item \textbf{Input:} Transformer sub block output $\bm{h}$ as a $\bm{T \times D}$ matrix, drop rate $p$, $mode$
\item \textbf{if} $mode == Inference$ \textbf{then}
\item \qquad return $\bm{h}$
\item \textbf{end if}
\item Randomly sample $\xi$: ${\xi _i} \sim {\text{Bernoulli}}(p)$, $i \leqslant D$
\item For each element ${\xi _i}$, create a mask vector ${M_i} \in {\mathbb{R}^T}$ with all elements equal to ${\xi _i}$.
\item Apply the mask: $\bm{h = h \times M}$
\item Normalization: $\bm{h} = \bm{h} \times count(\bm{M})/count\_ones(\bm{M})$
\end{enumerate}
\label{algorithm1}
\end{algorithm}

\section{Experimental Setup}
\subsection{Datasets}
To prove the effectiveness and generalization of DropDim, we conduct experiments on three types of sequence-to-sequence tasks: ASR, MT, and ST. MUST-C is a speech translation corpus, including triplets of speech, transcription, and translation, which can perform different tasks flexibly. It is built from English TED talks, covering 8 language pairs. We conduct experiments on the English-German language pair. There are 234K utterances totaling 408 hours in this parallel corpus, which is divided into training set (400 hours, 229,703 pronunciations), development set (3 hours, 1423 pronunciations), and tst-COMMON\footnote{MuST-C is a multilingual dataset and this testset is the commonly shared utterances between the languages.} set (5 hours, 2641 pronunciations).

\subsection{Preprocessing and Evaluation}
All the acoustic features used in this paper are 80-dimensional MFCC extracted using Kaldi \cite{povey2011the} with global cepstral mean and variance normalization. The text data in the source language is normalized with lowercase conversion, tokenization, and punctuation removal. We apply punctuation normalization instead of punctuation removal for the text data in the target language. The processing of the text is realized by Moses scripts\footnote{https://www.statmt.org/moses/}. We apply BPE \cite{sennrich2016neural} on the combination of source and target text to obtain shared subword units, and the vocabulary size is set to 8K.

The case-insensitive BLEU \cite{papineni2002bleu} calculated by the multi-bleu.pl script\footnote{https://github.com/moses-smt/mosesdecoder/scripts/generic/multi-bleu.perl} is used to evaluate translation tasks (MT, ST). Moreover, we use Word Error Rate (WER) to evaluate the ASR system.

\subsection{Training Settings}
The models involved in this paper all adopt the Transformer structure. We conduct experiments based on Espnet \cite{inaguma2020espnet} and use PyTorch-lightning \cite{falcon2019pytorch} to organize our code. For both encoder and decoder in the MT model, the number of layers is 6, the number of attention heads is 4, the embedding dimension is 256, and the filter size of the feed-forward neural network is 2048. The attention and residual dropout rates are 0.0 and 0.1, respectively. For the ST and ASR model, the hyperparameters of encoder and decoder are 12 and 6, respectively. Other parameters remain the same with MT model. We train our models with Adam optimizer \cite{kingma2015adam} on 4 NVIDIA V100 GPUs.

\section{Experimental Results}
\subsection{Main Results}

\begin{table}[th]
\centering
\caption{Model performance under different methods. *: The application location of dropout here is the same as DropDim.}
\label{tab:zhang.t1}
\begin{tabular}{l|c|c|c}
\hline
Models                                &ASR(WER\%$\bm{\downarrow}$) &MT(BLEU$\uparrow$)  &ST(BLEU$\uparrow$) \\
\hline
Transformer(base)                     &19.1                       &26.90               &21.13             \\
\hline
+dropout*                             &18.7          			  &27.10               &21.54             \\
+DropAttention                        &18.3                       &27.38               &21.34             \\
+Drophead                             &18.1                       &27.51               &21.57             \\
+label smoothing(0.1)                 &16.8                       &27.88               &21.31             \\
\hline
+Dropdim(random)                      &17.8                       &27.22               &21.85             \\
\quad +label smoothing(0.1)           &15.3                       &28.32               &22.59             \\
+Dropdim(span)                        &17.7                       &27.42               &21.97             \\
\quad +label smoothing(0.1)           &15.1                       &28.38               &22.99             \\
\hline
\end{tabular}
\end{table}

\begin{table}[th]
\centering
\caption{Experiments on different datasets and language pairs in ST task. Label smoothing is used in all methods by default.}
\label{tab:zhang.t2}
\begin{tabular}{l|c|c}
\hline
Models				&Librispeech En-Fr    &MUST-C En-Fr \\
\hline
base 				&16.08 				  &33.15 		\\
+Dropdim(random) 	&17.12 				  &34.46 		\\
+Dropdim(span)      &17.34 				  &34.57		\\
\hline
\end{tabular}
\end{table}

We first experimentally verify the effectiveness of DropDim for ASR, MT, and ST tasks, and compare it with other methods. The results are summarized in Table~\ref{tab:zhang.t1}.

{\bfseries DropDim vs Dropout.} The original transformer model did not apply dropout to the output of the residual connection. For the convenience of discussion, we define a transformer baseline in which DropDim in Equation~\ref{equa} is replaced with dropout. In all three tasks, DropDim and dropout share a similar trend and DropDim has a large gain compared to the dropout results. This shows evidence that the DropDim is a more effective regularizer compared to dropout.

{\bfseries Comparison with DropAttention and Drophead.} We compare with two popular structured dropout methods designed for self-attention mechanisms. All three methods can improve model performance. Specifically, on speech-related tasks (ASR, ST), DropDim achiceves better results. We believe this is because compared to DropAttention and Drophead, DropDim processes the input of the self-attention mechanism, which can completely drop semantic information and produce a stronger regularization effect. However, compared with speech, the text is coarse-grained, too strong regularization will hurt the MT model.

In the above comparison, DropAttention, Drophead, and dropout* do not use label smoothing. The same trend can be observed when label smoothing is added, and the results were not listed due to the limited number of pages.

{\bfseries Span vs Random.} DropDim(span) delivers better results on all three tasks. This is due to the fact that in the case of high dimensions (256 in this article), it may not be that a single embedding dimension determines a semantics, but several adjacent embedded dimensions jointly express a semantics. And semantic information can be removed more effectively with continuous drop, which brings greater challenges to model training. We leave the specific details for future research.

{\bfseries Integration with other regularization techniques.} Label smoothing \cite{szegedy2016rethinking} is a common regularization technique. The results in Table~\ref{tab:zhang.t1} show that both label smoothing and DropDim can improve model performance. Moreover, the performance can be further improved when combining DropDim and label smoothing. This shows that the overfitting problem is a direction that still needs to be studied, and there is no method that can completely solve it when used alone.

{\bfseries Results under different langugae pair and datasets.} In order to further proves the effectiveness and generality of Dropdim, we performed additional experiments on the Librispeech En-Fr \cite{kocabiyikoglu2018augmenting} and MUST-C En-Fr datasets in ST task. The results are listed in Table~\ref{tab:zhang.t2}. The experimental results showed the same trend as on MUST-C En-De datasets. The speech in Librispeech En-Fr was recorded on-site, which contains more noise, resulting the transcription and translation are not well aligned with the original audio. So, the model performance on this dataset is lower than on MUST-C En-Fr.

\subsection{Ablation Studies}
\begin{figure}
\centerline{\includegraphics[width=\columnwidth]{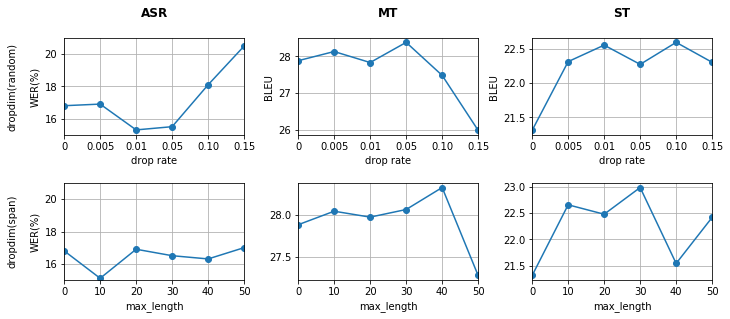}}
\caption{The effect of different hyperparameters on model performance.}
\label{fig:Hyperparameters}
\end{figure}

{\bfseries Effect of Hyperparameters.} The hyperparameters of Dropdim determine the number of embedding dimensions that are dropped. The influence of hyperparameters on model performance is summarized in Fig.~\ref{fig:Hyperparameters}. It can be observed that determining a sweet point of the drop ratio is critical to the generalization ability of the resulting model. Specifically, when DropDim (random) is used, the best drop ratio for ASR, MT, and ST are 0.01, 0.05, and 0.10, respectively. While using DropDim (span), the best max\_length ($\alpha$) is 10, 40, and 30, respectively. Dropping too many embedding dimensions will bring a very small improvement to the model and even hurt the performance. This may be because the model unable to extract knowledge from the input for learning. 

{\bfseries Effect on Different Part.} In this section, we study the effect of DropDim on different parts of the model. Specifically, we apply DropDim separately on the encoder, decoder, and the whole model. The results are shown in Table~\ref{tab:zhang.t3}. We can see that the proposed approach is more effective when used on the whole model because of the strong regularization effect.

\begin{table}[]
\centering
\caption{BLEU score of models when DropDim is applied on different part during training.}
\label{tab:zhang.t3}
\begin{tabular}{cccc}
\hline
\multirow{2}{*}{task} & \multicolumn{3}{c}{Part} \\ \cline{2-4}
                      & encoder  & decoder & all \\
\hline
ASR(WER\%$\bm{\downarrow}$)                     &16.9          &16.2         &15.1     \\
MT(BLEU$\uparrow$)                              &27.82         &27.71        &28.38     \\
ST(BLEU$\uparrow$)                              &22.66         &22.13        &22.99     \\
\hline
\end{tabular}
\end{table}

\subsection{Analysis}

\begin{figure}
\centerline{\includegraphics[width=2in]{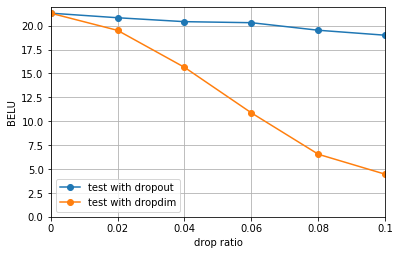}}
\caption{The performance of the model using dropout or DropDim during testing.}
\label{fig:semantic}
\end{figure}

{\bfseries DropDim drops more semantic information.} We first train the model normally on ST task and then apply dropout or DropDim when testing. For a fair comparison, DropDim (random) is used here because with the same drop rate, the same number of neurons are dropped. The difference is that neurons are randomly dropped when dropout is used while using DropDim (random) will drops the entire embedding dimension. It can be seen that as the drop ratio increases, the yellow curve in Fig.~\ref{fig:semantic} drops rapidly, indicating that DropDim can remove semantic information more effectively than dropout.

\begin{figure}
\centerline{\includegraphics[width=\columnwidth]{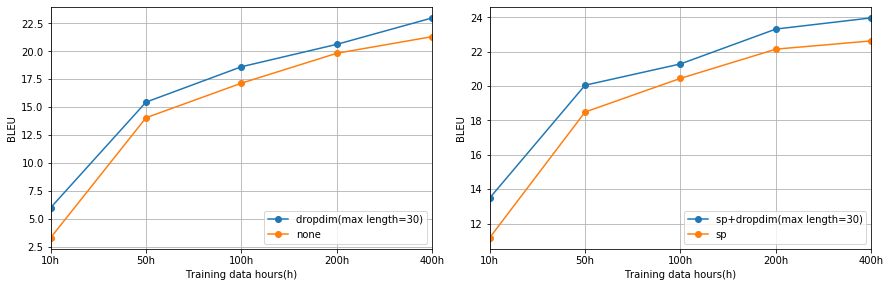}}
\caption{Model performance with or without DropDim under different amounts of data. Left: No data augmentation is used. Right: Speed Perturbs of 0.9, 1.0, 1.1 are used for data augmentation.}
\label{fig:amount}
\end{figure}

{\bfseries Results under different amounts of data.} We test the effectiveness of DropDim under different amounts of data on the ST task. To simulate different data resource scenarios, we randomly select 10-hour, 50-hour, and 100-hour speech training data from the MUST-C English-German dataset, which originally contains 400-hour speech. Then the model is trained with or without DropDim. In addition, Speed Perturb \cite{ko2015audio} is a commonly used data augmentation method under low resource conditions. We also study the combination of DropDim and Speed Perturb. The results are shown in Fig.~\ref{fig:amount}. The blue curve in the figure continues to be higher than the yellow curve, showing that DropDim can continuously improve the model performance under different amounts of data. Moreover, the combination with Speed Perturb can produce better results, indicating that the two methods are complementary.

\subsection{Visualization}
\begin{figure}
\centerline{\includegraphics[width=\columnwidth]{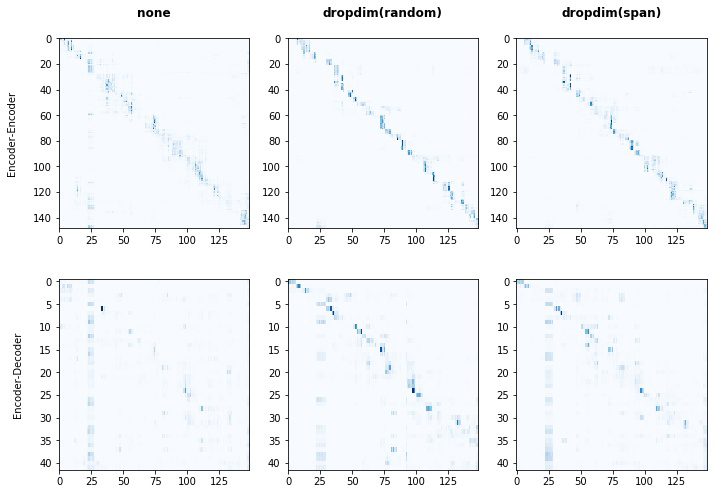}}
\caption{The visualization of attention of different modules under different strategies. The horizontal coordinates represent the sequence of speech frames. In the attention map of Encoder-Encoder, the vertical coordinates represent the sequence of speech frames. In the attention map of Encoder-Decoder, the vertical coordinates represent the text token sequence.}
\label{fig:attention}
\end{figure}

{\bfseries Attention Visualization.} We further analyze how DropDim improve the model performance by visualizing the encoder-encoder and encoder-decoder attention maps. Specifically, the encoder-encoder attention is extracted from the last layer of the encoder, and the encoder-decoder attention is extracted from the last layer of the decoder. Fig.~\ref{fig:attention} shows an example. The attention in the ST model tends to be smooth between input frames. However, after applying DropDim, both types of attention become more concentrated, indicating that DropDim can not only help the model to remove interference but also make the model better align with the target.

\begin{figure}[th]
\centerline{\includegraphics[width=2in]{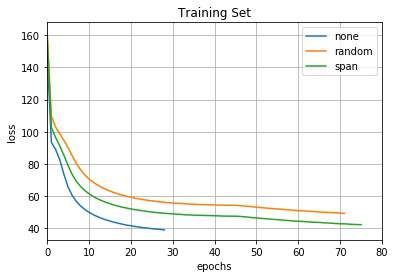}}
\caption{Training process under different strategies.}
\label{fig:train}
\end{figure}

{\bfseries Training Process Visualization.} It can be seen from the network training curve in Fig.~\ref{fig:train} that when DropDim is used, the network can underfit the training loss, which is in sharp contrast with the usual situation where networks tend to over-fit to the training data. This is consistent with specaugment \cite{park2019specaugment}, which converts over-fitting problems into under-fitting problems, and improves model performance through longer training time.

This also reveals to us that the combination of Dropdim and specaugment is not suitable. Because although this combination can improve performance, the training time will be further increased. Dropdim is more suitable to be combined with Speed Perturbs, because the latter improves the model performance by increasing the amount of data, which is different from Dropdim and specaugment.

\section{Conclusion}
In this paper, we introduce DropDim, a simple regularization method for training transformer models. We have verified the effectiveness of DropDim on a wide range of tasks, showing that it can continuously improve the model performance. Experimental analysis show that DropDim is also competitive under low-resource conditions and is complementary to other data augmentation methods.
\bibliographystyle{IEEEtran}

\bibliography{mybib}

\end{document}